\documentclass[journal]{IEEEtran}

\usepackage{amsmath}
\usepackage{amssymb}
\usepackage{todonotes}
\usepackage{multirow}
\usepackage{verbatim}
\usepackage{array}
\usepackage{multicol}
\usepackage{xcolor,colortbl}
\usepackage{tabu}
\usepackage{hyperref}

\newcommand\xuparrow[1]{%
\ensuremath{
   \mathrel{\rotatebox{270}{$\xleftarrow{\rule{#1}{0pt}}$}}
   }
}
\newcolumntype{K}[1]{>{\centering\arraybackslash}p{#1}}
\newcommand{\vh}{\ensuremath{\mathbf{h}}}

\begin{document}
\title{High-Resolution Breast Cancer Screening with Multi-View Deep Convolutional Neural Networks}

\author{Krzysztof J. Geras$^{1, 3}$, Stacey Wolfson$^{3}$, Yiqiu Shen$^{1}$, Nan Wu$^{1}$, S. Gene Kim$^{3, 4}$, Eric Kim$^{3}$,\\Laura Heacock$^{3}$, Ujas Parikh$^{3}$, Linda Moy$^{3, 4}$, Kyunghyun Cho$^{1, 2, 5}$

\thanks{$^{1}$Center for Data Science, New York University, 60 5th Ave, New York, NY 10011}
\thanks{$^{2}$Courant Institute of Mathematical Sciences, New York University, 251 Mercer St, New York, NY 10012}
\thanks{$^{3}$Center for Biomedical Imaging, Radiology, NYU School of Medicine, 660 1st Avenue, New York, NY 10016}
\thanks{$^{4}$Perlmutter Cancer Center, NYU Langone Medical Center, 160 E 34th St, New York, NY 10016}
\thanks{$^{5}$CIFAR Azrieli Global Scholar}
\thanks{\tiny{This work has been submitted to the IEEE for possible publication. Copyright may be transferred without notice, after which this version may no longer be accessible.}}
}

\maketitle

\begin{abstract}
Advances in deep learning for natural images have prompted a surge of interest in applying similar techniques to medical images. The majority of the initial attempts focused on replacing the input of a deep convolutional neural network with a medical image, which does not take into consideration the fundamental differences between these two types of images. Specifically, fine details are necessary for detection in medical images, unlike in natural images where coarse structures matter most. This difference makes it inadequate to use the existing network architectures developed for natural images, because they work on heavily downscaled images to reduce the memory requirements. This hides details necessary to make accurate predictions. Additionally, a single exam in medical imaging often comes with a set of views which must be fused in order to reach a correct conclusion. In our work, we propose to use a multi-view deep convolutional neural network that handles a set of high-resolution medical images. We evaluate it on large-scale mammography-based breast cancer screening (BI-RADS prediction) using 886,000 images. We focus on investigating the impact of the training set size and image size on the prediction accuracy. Our results highlight that performance increases with the size of training set, and that the best performance can only be achieved using the original resolution. In the reader study, performed on a random subset of the test set, we confirmed the efficacy of our model, which achieved performance comparable to a committee of radiologists when presented with the same data.
\end{abstract}


\IEEEpeerreviewmaketitle

\section{Introduction}

Breast cancer is the second leading cancer-related cause of death among women in the United States. It is estimated that 232,000 women were diagnosed with breast cancer and approximately 40,000 died from the disease in 2015 \cite{RN53}. Screening mammography is the main imaging test used to detect occult breast cancer. Multiple randomized clinical trials have shown a 30\% reduction in mortality in asymptomatic women who were undergoing screening mammography \cite{RN38, RN40}. Although mammography is the only imaging test that reduced breast cancer mortality \cite{RN38, RN40, RN41}, the appropriate screening interval for mammograms has been the subject of public debate with different professional societies offering varying guidelines for mammographic screening \cite{RN38, RN40, RN41, RN37}. In particular, there has been public discussion regarding the potential harms of screening. These harms include false positive recalls and false positive biopsies as well as anxiety caused by recall for diagnostic testing after a screening exam. Overall, the recall rate following a screening mammogram is between 10-15\%. This equates to about 3.3 to 4.5 million callback exams for additional testing \cite{RN54}.

The vast majority of the women asked to return following an inconclusive mammogram undergo another mammogram and/or ultrasound for clarification. Most of these false positive findings are found to represent normal breast tissue. Only 10\% to 20\% of women who have an abnormal screening mammogram are recommended to undergo a biopsy. Only 20-40\% of these biopsies yield a diagnosis of cancer \cite{RN43}. In 2014, over 39 million screening and diagnostic mammography exams were performed in the US. Therefore, in addition to the anxiety from undergoing a false positive mammogram, there are significant costs associated with unnecessary follow ups and biopsies. Clearly, there is an unmet need to shift the balance of routine breast cancer screening towards more benefit and less harm.

\subsection{Breast Cancer Screening as a Deep Learning Task}

Deep learning has recently seen enormous success in challenging problems such as object recognition in natural images, automatic speech recognition and machine translation \cite{lecun2015deep}. This success has prompted a surge of interest in applying deep convolutional networks (DCN) to medical imaging. Many recent studies have shown the potential of applying such networks to medical imaging, including breast screening mammography; however, without investigating the fundamental differences between medical and natural images and their impact on the design choices and performance of proposed models. For instance, many recent works have either significantly downscaled a whole image or focused on classifying a small region of interest. This might be detrimental to performance of such models given the well-known dependency of breast cancer screening on fine details, such as the existence of a cluster of microcalcifications, as well as global structures, for example the symmetry between two breasts. Furthermore, the potential of DCNs has only been assessed in limited settings of small data sets often consisting of less than one thousand images, while the success of such networks in natural object recognition is largely attributed to the availability of more than one million annotated images. This further hinders our understanding of the true potential of DCNs in medical imaging, particularly in breast cancer screening.

In this work, we conducted an investigation into analyzing and understanding fundamental properties of deep convolutional networks in the context of breast cancer screening. We started by building a large-scale data set of 201,698 screening mammographic exams (886,437 images) collected at multiple sites of our institution. We developed a novel DCN that is able to handle multiple views of screening mammography and to utilize large high-resolution images without downscaling. We refer to this DCN as a multi-view deep convolutional network (MV-DCN). Our network learns to predict the assessment of a radiologist, classifying an incoming example as BI-RADS 0 (``incomplete''), BI-RADS 1 (``normal'') or BI-RADS 2 (``benign finding''). We studied the impact of the data set size and image resolution on the screening performance of the proposed MV-DCN, which would serve as a \textit{de facto} guideline for optimizing future deep neural networks for medical imaging. We further investigated the potential of the proposed MV-DCN by visualizing its predictions. Finally, we conducted a reader study, which showed that our model, on a random subset of the test set, is almost as accurate as a committee of radiologists presented with the same data. Furthermore, we found that we obtain the best results by averaging the predictions of our model's with the predictions of the committee of the radiologist.

\section{High-Resolution Multi-View Deep Convolutional Neural Networks}

\subsection{Deep Convolutional Neural Network} 

A deep convolutional neural network \cite{lecun1989backpropagation, lecun1998gradient} is a classifier that takes an image $\mathbf{x}$ as input, often with multiple channels corresponding to different colors (e.g., RGB), and outputs the conditional probability distribution over the categories $p(y|\mathbf{x})$. This is done by a series of nonlinear functions that gradually transform the input pixel-level image. A major property of the deep convolutional network, which distinguishes it from a multi-layer perceptron, is that it heavily relies on convolutional and pooling layers, which make the network invariant to local translation of visual features in the input.

\subsection{Multi-View Deep Convolutional Neural Network}

\begin{figure*}
\begin{center}
\begin{tabular}{c c c c}
     \multicolumn{4}{c}{\footnotesize{(A) BI-RADS 0}}\vspace{-0.5mm}\\
     \includegraphics[scale=0.4, trim=1mm 0mm 1mm 0mm]{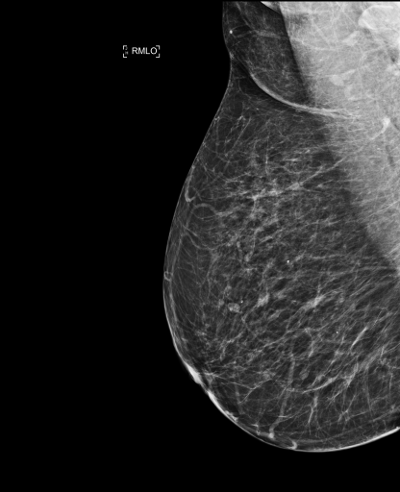}&
     \includegraphics[scale=0.4, trim=1mm 0mm 1mm 0mm]{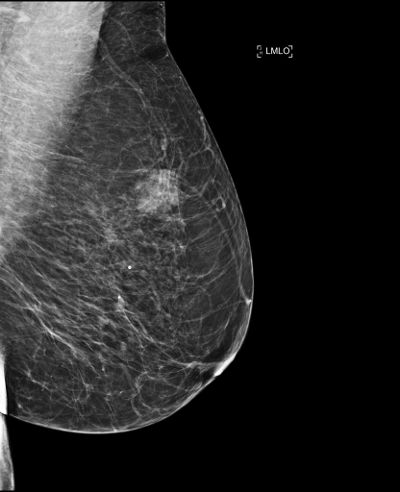}&
     \includegraphics[scale=0.4, trim=1mm 0mm 1mm 0mm]{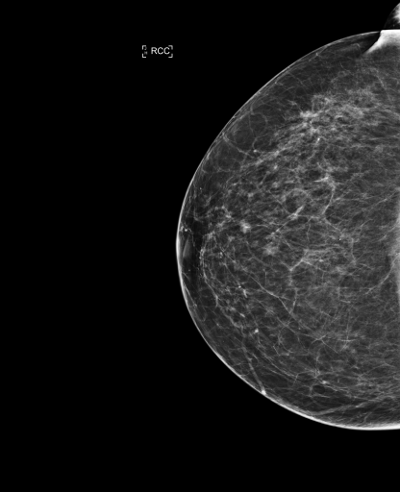}&
	 \includegraphics[scale=0.4, trim=1mm 0mm 1mm 0mm]{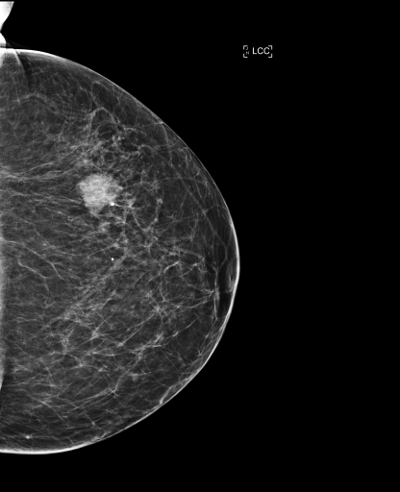}\\
     \vspace{-4mm}\\
     \multicolumn{4}{c}{\footnotesize{(B) BI-RADS 1}}\vspace{-0.5mm}\\
     \includegraphics[scale=0.4, trim=1mm 0mm 1mm 0mm]{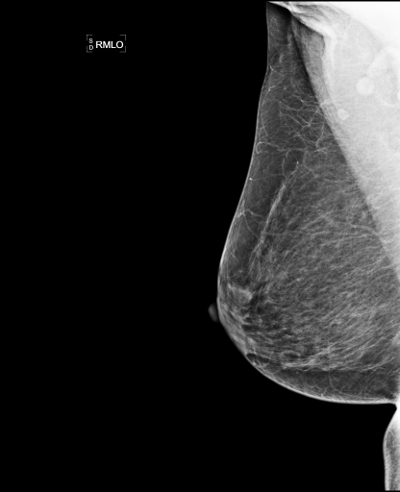}&
     \includegraphics[scale=0.4, trim=1mm 0mm 1mm 0mm]{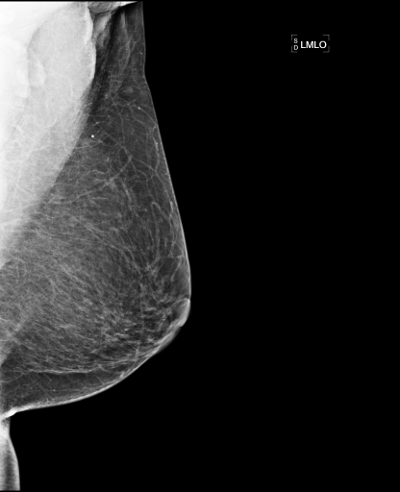}&
     \includegraphics[scale=0.4, trim=1mm 0mm 1mm 0mm]{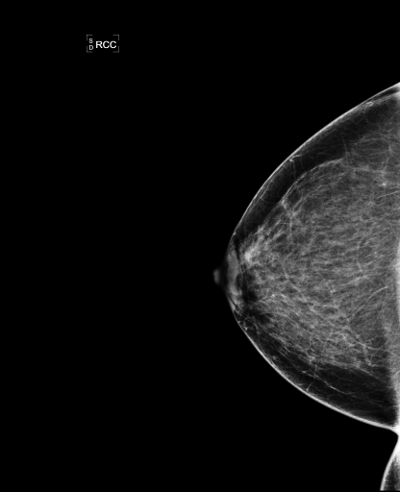}&
     \includegraphics[scale=0.4, trim=1mm 0mm 1mm 0mm]{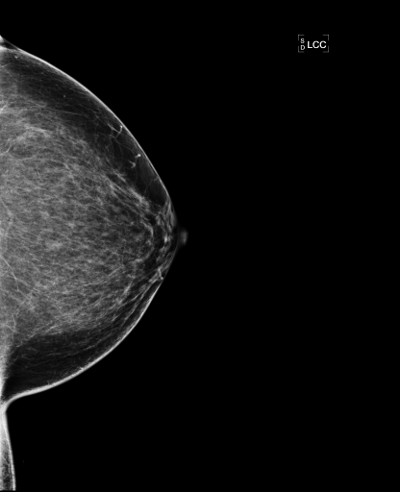}\\
     \vspace{-4mm}\\
     \multicolumn{4}{c}{\footnotesize{(C) BI-RADS 2}}\vspace{-0.5mm}\\
     \includegraphics[scale=0.4, trim=1mm 0mm 1mm 0mm]{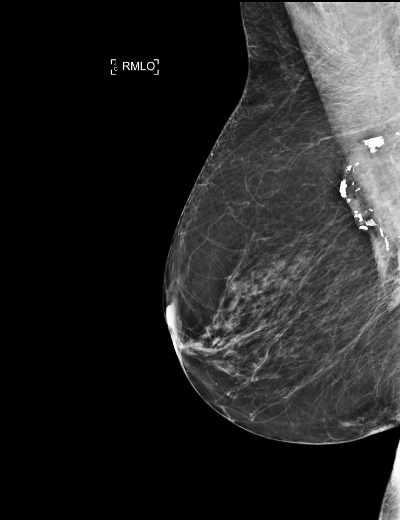}&
     \includegraphics[scale=0.4, trim=1mm 0mm 1mm 0mm]{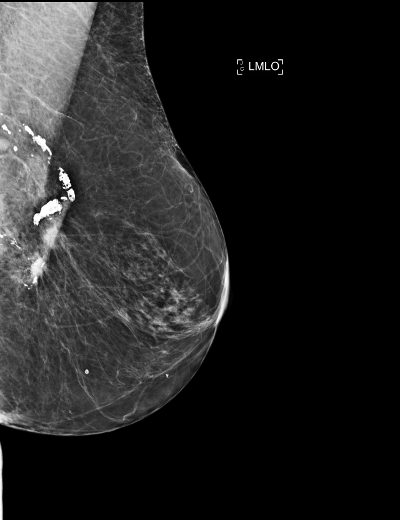}&
     \includegraphics[scale=0.4, trim=1mm 0mm 1mm 0mm]{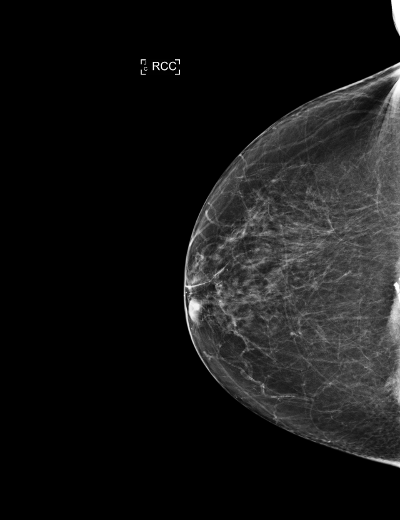}&
	 \includegraphics[scale=0.4, trim=1mm 0mm 1mm 0mm]{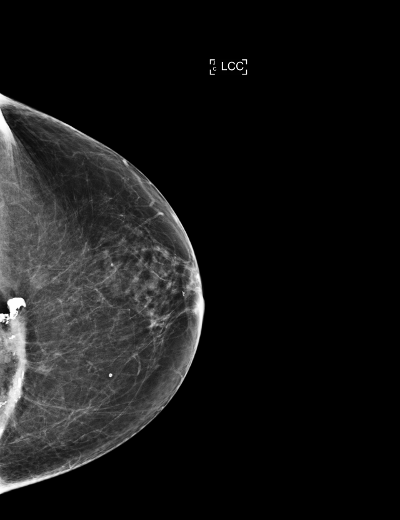}\\
     \footnotesize{\shortstack{right mediolateral olique\\(R-MLO)}} &
     \footnotesize{\shortstack{left mediolateral olique\\(L-MLO)}} & 
     \footnotesize{\shortstack{right cranio-caudal\\(R-CC)}} & 
	 \footnotesize{\shortstack{left cranio-caudal\\(L-CC)}}
\end{tabular}
\end{center}
\vspace{-2mm}
\caption{\label{fig:views}The four standard views used in our experiments for exams categorized as BI-RADS 0 (A), BI-RADS 1 (B) and BI-RADS 2 (C). (A) In the left breast, there is a round mass with irregular margins. The patient was recalled, because additional mammographic views and a breast ultrasound were necessary to further characterize this mass. This mass turned out to be an invasive ductal carcinoma. (B) This is a normal mammogram in a patient with a scattered fibroglandular tissue breast pattern. No abnormalities were seen. (C) This patient has a scattered fibroglandular tissue breast pattern. In the posterior depth of the both breasts, near the chest wall, there are calcified masses consistent with post-surgical changes.}
\end{figure*}

Object recognition tasks with natural images usually involve only one object at a time, in contrast an exam in medical imaging often comes with a set of views. For instance, it is standard in screening mammography to obtain cranial caudal (CC) and mediolateral oblique (MLO) views for each breast of a patient, resulting in a set of four images. We will refer to them as L-CC, R-CC, L-MLO and R-MLO (Figure \ref{fig:views}).

There is a rich literature on building deep neural networks for multi-view data. Most of them fall into one of two families. First, there are works on unsupervised feature extraction from multiple views using a variant of deep autoencoders \cite{ngiam2011multimodal, srivastava2012multimodal, wang2015deep}. They usually train a multi-view deep neural network with unlabeled examples, and use the output of such a network as a feature extractor, followed by a standard classifier. On the other hand, Su~et~al.~\cite{su2015multi} proposed to build a multi-view deep convolutional network directly for classification.

We propose a variant of MV-DCN which was motivated by Su~et~al.~\cite{su2015multi}. This MV-DCN computes the output in two stages. In the first stage, a number of convolutional and pooling layers is separately applied to each of the views. We denote such view-specific representation by $\vh_\mathrm{v}$, where $\mathrm{v}$ refers to the index of the view. These view-specific representations are concatenated to form a vector, $\left[\vh_\mathrm{L-CC}, \vh_\mathrm{R-CC}, \vh_\mathrm{L-MLO}, \vh_\mathrm{R-MLO}\right]$, which is an input to the second stage - a fully connected layer followed by a softmax layer producing output distribution $p(\mathbf{y}|x)$.

The whole network is trained jointly by stochastic gradient descent with backpropagation \cite{rumelhart1986learning}. Furthermore, we employ a number of regularization techniques to avoid the behavior of overfitting due to the relatively small size of training dataset, such as data augmentation by random cropping \cite{krizhevsky2012imagenet} and dropout \cite{srivastava2014dropout}. These will be described later in detail.

\subsection{High-Resolution Convolutional Neural Network}

It is common in object recognition and detection in natural images to heavily downscale an original high-resolution image. For instance, the input to the network of the best performer in ImageNet Challenge 2015 (classification task) was an image downscaled to $224 \times 224$ \cite{he2015deep}. This is often done to improve the computational efficiency, both in terms of computation and memory, and also because no significant improvement has been observed with higher-resolution images. It reflects an inherent property of natural images, in which the objects of interest are usually presented in relatively larger portions than other objects and what matters most are their macro-structures, such as shapes, colors and other global descriptors. However, downscaling of an input image is not desirable in the case of medical images, and in particular for early-stage screening based on breast mammography. Often a cue for diagnosis is a subtle finding which may be identified only at the original resolution.

In order to address the computational issues of handling full-resolution images, we propose to use aggressive convolution and pooling layers. First, we use convolution layers with strides larger than one in the first two convolutional layers. Also, the first pooling layer has a larger stride than the other pooling layers. As a result of this, we greatly reduce the size of feature maps early in the network. Although this aggressive convolution and pooling loses some spatial information, the parameters of the network are adjusted to minimize this information loss during training. This is unlike downscaling of the input, which loses information unconditionally. Second, we average feature maps in the last layer before concatenating them \cite{network_in_network}, instead of simply flattening the feature maps and then concatenating them \cite{krizhevsky2012imagenet, vgg}. This drastically reduces the dimensionality of the view-specific vector without much, if any, performance degradation \cite{szegedy2015going}. Using both of these approaches, we are able to build an MV-DCN that takes four $2600 \times 2000$ pixels images (one per view) as input without any downscaling.

\section{Related Work}

Let us briefly review recent deep learning based approaches to breast mammography, summarized in Table~\ref{tab:refs}.

\paragraph{Multi-Stage vs. End-to-End Approaches.}

Traditionally breast cancer screening and lesion detection are done in three stages: detection, analysis and final assessment/management. In the first stage, a breast mammography image is segmented into different types of regions, such as foreground (breast) and background. Within the segmented region of breast, the second stage focuses on extracting a set of regions of interest (ROI) that will be examined in more detail. In the third stage, each of those ROI's is determined to be a malignant lesion or not. The outcome of the third stage is used to make the final decision on a given case consisting of multiple views.

Most of the recent research on applying deep learning to breast mammography has focused on replacing one or more stages in this existing multi-stage pipeline; for instance, mass detection \cite{Domingues2013MassDO, dhungel2015automated, ertosun2015probabilistic}. In their work, a deep neural network is trained to determine whether a small patch is a mass. Others have focused on training a deep neural network for classifying a small region of interest into one of a few categories, assuming an existing mass detection system \cite{huynh2016digital,levy2016breast,arevalo2016representation,mordang2016automatic}. 

On the other hand, a small number of research groups have considered replacing the whole multi-stage approach with a single, or a series of, trainable machine learning algorithms. Kooi~et~al.~\cite{kooi2017large} proposed to use a random forest classifier for mass detection followed by a DCN that classifies each detected mass. A similar approach was proposed by Becker~et~al.~\cite{becker2017deep}. Akselrod-Ballin~et~al.~\cite{akselrod2016region} further proposed to use deep convolutional networks for both mass detection and classification, potentially enabling end-to-end training. Two groups \cite{zhu2016deep,carneiro2015unregistered} went even further by proposing a single deep convolutional network that classifies a whole image, or a set of multiple views. The work by Carneiro~et~al.~\cite{carneiro2015unregistered} is closest to our approach in this paper. In both works, a single deep convolutional neural network takes as input a set of multiple views of an exam and predicts its BI-RADS label.

\paragraph{Data Size.}

Although it is recognized that one of the driving forces behind the success of deep learning is the availability of large scale data, it has not been exploited when applying deep learning to mammography. As evident in Table~\ref{tab:refs}, most of the recent works use less than 1,000 images for both training and testing. To avoid the issue of small training data, most of the earlier works resorted to training with many small patches, or ROI's, avoiding end-to-end training. One exception is the work of Carneiro~et~al.~\cite{carneiro2015unregistered} in which they use the whole image with, however, the deep convolutional network pretrained for object recognition in natural images. Unlike these earlier approaches, we use a large-scale data set of an unprecedented size, consisting of 886,437 images. This allows us to carefully study the impact of the size of training data set.

\paragraph{Natural vs. Controlled Distribution.}

Breast screening is aimed at a general population rather than a selected group of patients. This implies that the distribution of the screening outcome is heavily skewed toward ``normal'' (BI-RADS 1). In our training set which closely follows a general population distribution, approximately 46\% of the cases were assigned BI-RADS 1 (``normal''), while 41\% were assigned BI-RADS 2 (``benign finding'') and 13\% BI-RADS 0 (``incomplete''). This is in contrast to two widely-used, publicly available datasets, \mbox{INBreast}~\cite{moreira2012inbreast}, DDSM~\cite{bowyer1996digital,heath1998current} and other curated small-scale datasets from recent literature (see those in Table~\ref{tab:refs}). These datasets are often constructed to include approximately the same proportions of normal and abnormal cases, resulting in, what we refer to as, a {\it controlled distribution} of outcomes which differs from a {\it natural distribution}. For instance, INBreast has approximately achieved a balance between benign and malignant cases. This type of artificial balancing, or equivalently upsampling of malignant cases, may bias a model to more often predict a given case as malignant and require a recall more often than necessary. Unlike these earlier works, in this paper, we use the full data without artificial balancing of outcomes to ensure that any trained deep convolutional network will closely reflect the natural distribution of outcomes. 

\begin{table}
\caption{\label{tab:refs} Previous works on deep learning for breast mammography.}
\begin{center}
\begin{tabular}{K{0.3cm} K{0.3cm} K{0.3cm} K{1.8cm} K{1.2cm} K{0.3cm} K{0.4cm} K{0.3cm}}
task$^\square$ & ref. & E2E$^\bullet$ & \#images$^\dagger$ & image size$^\ddagger$ & MV$^\heartsuit$ & input$^\spadesuit$ & dist.$^\circ$  \\
\hline\hline
\multirow{4}{*}{\rotatebox[origin=c]{90}{BI-RADS}} & $\star$ & $\surd$ & 829k (57k) & 2600$\times$2000 & $\surd$ & IMG & N \\
& \cite{carneiro2015unregistered} & $\surd$ & 680 ($\approx$ 340) & 264$\times$264 & $\surd$ & IMG & C \\
& \cite{zhu2016deep} & $\surd$ & 410 (CV) & 224$\times$224 & $\surd$ &IMG & C \\
& \cite{akselrod2016region} & $\surd$ & 850 ($\approx 170$) & 800$\times$800 & & IMG & C \\
\hline
\multirow{5}{*}{\rotatebox[origin=c]{90}{lesion}} & \cite{huynh2016digital} &  & 607 (CV) & 512$\times$512 & & ROI & C \\
& \cite{kooi2017large} &  & 44,000 (18,000) & 250$\times$250 & & ROI & N  \\
& \cite{levy2016breast} &  & 1820 (182) & 224$\times$224 & & ROI & C \\
& \cite{arevalo2016representation} &  & 736 ($\approx$300) & 150$\times$150 & & ROI & C \\
& \cite{mordang2016automatic} &  & 1606 ($\approx$378) & 13$\times$13 & $\surd$ & ROI & N \\
\hline
\multirow{3}{*}{\rotatebox[origin=c]{90}{mass}} & \cite{Domingues2013MassDO} &  & 116 (CV) & 32$\times$32 & & ROI & C \\
& \cite{dhungel2015automated} &  & 410 (CV) &  264$\times$264 & & ROI & C \\
& \cite{ertosun2015probabilistic} & $\surd$ &2500 (250) & 256$\times$256 & & ROI & C \\
\hline
\multirow{2}{*}{\rotatebox[origin=c]{90}{MC}} & \cite{wang2016discrimination} &  & 1000 (204) & N/A$^\clubsuit$ & & ROI & C \\
& \cite{bekkermulti} &  & 1410 (N/A) & N/A$^\clubsuit$ & & ROI & C \\
\end{tabular}
\end{center}

When more than one data set was used, we list the size of the largest data set. $\star$~denotes this paper. The table should be read with the following footnotes. $\square$~The target task; BI-RADS: BI-RADS prediction, lesion: lesion classification (benign vs. malignant), mass: mass detection, and MC: microcalcification detection.
$\bullet$~Whether the proposed system is trainable end-to-end. For instance, a system that requires an external system for extracting regions of interest (ROI) is not end-to-end, while a system that uses convolutional networks for both ROI extraction and lesion classification is. 
$\dagger$~In the parentheses there is the number of test images or ``CV'' if cross-validation is used. 
$\ddagger$ denotes the size of the input image to a deep neural network.
$\heartsuit$~Whether multiple views per one exam are utilized.
$\circ$~Whether the data reflects natural distribution (N) or controlled distribution (C). 
$\spadesuit$~Whether the input to a deep neural network is a whole image (IMG) or a small subset (ROI). $\clubsuit$~Did not use images as input to the learning algorithm.
\end{table}

\section{Data}

\subsection{Collection}

This is a Health Insurance Portability and Accountability (HIPAA)-compliant, retrospective study approved by our Institutional Review Board. Consecutive screening mammograms for 129,208 patients aged\footnote{When more than one exam for a patient was in the data set, we included the ages of that patient at the time of all exams to compute the values above.} between 19 and 99 (mean: 57.2, std: 11.6) collected within seven years (2010-2016) at five imaging sites affiliated with New York University School of Medicine were used in this study. These imaging centers are located in the New York City metropolitan area (a large academic center and two large ambulatory care practices), where, altogether, over 70,000 mammograms are performed annually. The ethnic makeup of the patient cohort for this study reflects the population pool in NYC, which is 50\% Caucasian, 30\% African American, 5\% Asian and 15\% Hispanic.

\subsection{Data Statistics}

We used all data that we were able to collect and did not exclude any data unless they were acquired incorrectly\footnote{We only excluded images if they were of views which should not be taken in screening mammography, if they were technically not correct (e.g. if they did not have a time stamp), if they were smaller than $2600 \times 2000$ pixels in either of the corresponding dimensions or if their magnification factor was smaller than 1.0 or bigger than 1.1. We used all exams that had at least one correct image for each of the standard views.}. We divided the data into disjoint training, validation and test sets in the following manner. To start with we sorted all patients according to the date of their latest exam in the data set. We use the first 80\% of the patients in this order as the training data, the next 10\% as the validation data and the last 10\% as the test data. For each patient in the test set, we evaluate our model's performance in predicting only the label for the latest exam of each patient. This way we can reliably estimate the level of accuracy we would achieve if we tested our model on future exams. There are altogether 129,208 patients, 201,698 exams and 886,437 images in the data set.

\begin{table}[h!]
\centering
\caption{\label{tab:data_statistics} Distribution of data associated with different BI-RADS in training, validation and test data. Each cell in the table has the following format: number of exams / number of images.}

\begin{tabular}{ c | c c c}
& BI-RADS 0 & BI-RADS 1 & BI-RADS 2\\ \hline\hline
Training & 21946 / 95471 & 74832 / 327035 & 67446 / 298680 \\
Validation & 2634 / 11471 & 11542 / 50627 & 10376 / 46178 \\
Test & 1341 / 5871 & 5986 / 26213 & 5595 / 24891 \\
\end{tabular}
\end{table}

\subsection{Data preprocessing and augmentation}
\label{sec:preprocessing}

We normalized the images in the following way. For each image we computed the mean, $\mu$, and the standard deviation, $\sigma$, of its pixels. We then subtracted $\mu$ from each pixel and divided each pixel by $\sigma$. Additionally, we flipped horizontally the images of R-CC and R-MLO views so that the breast was always on the same side of the image. 

Since the images vary in size and a large fraction of the surface of each image is empty, we cropped all of them to the size of $2600 \times 2000$ pixels. We did it for two reasons. First, to unify the sizes of the images (which we need to put them in mini-batches during training) while keeping them at a similar scale and, second, to avoid processing the background which does not contain any information. 
The position of the crop was determined in the following manner. First, the crop area was placed leftmost on the horizontal axis and centrally on the vertical axis. To augment the data set, noise was added to this position. Let us denote the number of pixels between the top border of the crop area and the top border of the image by $b_\text{top}$ and analogously define $b_\text{bottom}$ and $b_\text{right}$. We drew a number, $t_\text{vertical}$ from a uniform distribution $\mathcal{U}(-\mathrm{min}(b_\text{top}, 100), \mathrm{min}(b_\text{bottom}, 100))$ and $t_\text{horizontal}$ from $\mathcal{U}(0, \mathrm{min}(b_\text{right}, 100))$. Finally we translated the position of the crop area by $t_\text{horizontal}$ pixels horizontally and $t_\text{vertical}$ pixels vertically. During training this noise was sampled independently every time an image is used.
During validation there was no noise added to the position of the crop area.
At test time, we fed ten sets of four randomly cropped views to the network. The final prediction was made by averaging predictions for all crops. The aim of this averaging is twofold; first, to use information from outside the center of the image while keeping the size of the input fixed and second, to make prediction of the network more stable. A small fraction of data contains more than one image per view. For such cases one image per view was sampled randomly and uniformly each time an exam was used during training and testing. During validation the image with the earliest time stamp was always used.

\section{Settings}

\subsection{Evaluation Metrics}

When there are two classes the most frequently applied performance metric is the AUC (area under the ROC curve). However, since there are three classes in our learning task, we cannot apply this metric directly. Instead we computed three AUCs, each time treating one of the three classes as a positive class and the remaining two as negative. We used the macro average of the three AUCs, abbreviated as macAUC, as the main performance metric in this work. 

Unlike other widely used nonlinear classifiers, such as a support vector machine or a random forest, a deep convolutional neural network outputs a proper conditional distribution $p(y|\mathbf{x})$. It allows us to compute the network's confidence in its prediction by computing the entropy of this distribution, i.e.,
\begin{align}
H(y|\mathbf{x}) = -\sum_{y' \in \mathcal{C}} p(y'|\mathbf{x}) \log p(y'|\mathbf{x}),
\label{eq:entropy}
\end{align}
where $y'$ iterates over all possible classes $\mathcal{C}$. The larger the entropy, the less confident the network is about its prediction. Based on $H$, we can quantify the change in accuracy (measured by AUC) with respect to the network's confidence.

\subsection{Model Setup}
\label{sec:setup}
\begin{figure}

\begin{center}
\small
\begin{tabular}{| c || c || c || c || c | c}
    \cline{1-1} \cline{2-2} \cline{3-3} \cline{4-4} \cline{5-5}
    \textbf{layer} & \textbf{kernel size} & \textbf{stride} & \textbf{\#maps} & \textbf{repetition} & \\
    \cline{1-1} \cline{2-2} \cline{3-3} \cline{4-4} \cline{5-5}
    \multicolumn{6}{c}{} \\[-7pt]
    
    \cline{1-3} \cline{4-4}
    \multicolumn{3}{|c||}{global average pooling} & \multicolumn{1}{c||}{256} & \multicolumn{2}{c}{}\\
    \cline{1-3} \cline{4-4}
    \multicolumn{5}{c}{} \\[-7pt]
    \cline{1-1} \cline{2-2} \cline{3-3} \cline{4-4} \cline{5-5}
    convolution & 3$\times$3 & 1$\times$1 & 256 & $\times$3 \\ 
    \cline{1-1} \cline{2-2} \cline{3-3} \cline{4-4} \cline{5-5}
    \multicolumn{5}{c}{} \\[-7pt]

    \cline{1-1} \cline{2-2} \cline{3-3} \cline{4-4} 
    max pooling & 2$\times$2 & 2$\times$2 & 128 \\ 
    \cline{1-1} \cline{2-2} \cline{3-3} \cline{4-4} \cline{5-5}
    convolution & 3$\times$3 & 1$\times$1 & 128 & $\times$ 3\\
    \cline{1-1} \cline{2-2} \cline{3-3} \cline{4-4} \cline{5-5}
    \multicolumn{5}{c}{} \\[-7pt]

    \cline{1-1} \cline{2-2} \cline{3-3} \cline{4-4} 
    max pooling & 2$\times$2 & 2$\times$2 & 128 \\ 
    \cline{1-1} \cline{2-2} \cline{3-3} \cline{4-4} \cline{5-5}
    convolution & 3$\times$3 & 1$\times$1 & 128 & $\times$ 3\\
    \cline{1-1} \cline{2-2} \cline{3-3} \cline{4-4} \cline{5-5}
    \multicolumn{5}{c}{} \\[-7pt]
    
    \cline{1-1} \cline{2-2} \cline{3-3} \cline{4-4} 
    max pooling & 2$\times$2 & 2$\times$2 & 64 \\ 
    \cline{1-1} \cline{2-2} \cline{3-3} \cline{4-4} \cline{5-5}
    convolution & 3$\times$3 & 1$\times$1 & 64 & $\times$ 2\\ 
    \cline{1-1} \cline{2-2} \cline{3-3} \cline{4-4} \cline{5-5}
    convolution & 3$\times$3 & 2$\times$2 & 64 \\ 
    \cline{1-1} \cline{2-2} \cline{3-3} \cline{4-4} 
    \multicolumn{5}{c}{} \\[-7pt]

    \cline{1-1} \cline{2-2} \cline{3-3} \cline{4-4} 
    max pooling & 3$\times3$ & 3$\times$3 & 32 \\ 
    \cline{1-1} \cline{2-2} \cline{3-3} \cline{4-4} 
    convolution & 3$\times$3 & 2$\times$2 & 32 \\ 
    \cline{1-1} \cline{2-2} \cline{3-3} \cline{4-4} 
    \multicolumn{5}{c}{} \\[-7pt]
    
    \cline{1-3} \cline{4-4} 
    \multicolumn{3}{|c||}{input} &  \multicolumn{1}{c||}{1} \\ 
    \cline{1-3} \cline{4-4} 
\end{tabular}
\end{center}
\caption{\label{fig:column} Description of one deep convolutional network column for a single view. It transforms the input view (a gray-scale image) into a 256-dimensional vector.}
\end{figure}

\begin{figure}
\centering
\small
\begin{tabular}{r | K{1.2cm} || K{1.2cm} || K{1.2cm} || K{1.2cm} |}
\cline{2-5}
\multirow{4}{*}{\xuparrow{12ex}}
& \multicolumn{4}{c|}{Classifier $p(y|x)$} \\
\cline{2-5}
\multicolumn{1}{c}{} & \multicolumn{4}{c}{} \\[-7pt]
\cline{2-5}
& \multicolumn{4}{c|}{Fully connected layer (1024 hidden units)} \\
\cline{2-5}
\multicolumn{1}{c}{} & \multicolumn{4}{c}{} \\[-7pt]
\cline{2-5}
& \multicolumn{4}{c|}{Concatenation (256$\times$4 dim)} \\
\cline{2-5}
\multicolumn{1}{c}{} & \multicolumn{4}{c}{} \\[-7pt]
\cline{2-2} \cline{3-3} \cline{4-4} \cline{5-5}
& DCN & DCN & DCN & DCN \\
\cline{2-2} \cline{3-3} \cline{4-4} \cline{5-5}
\multicolumn{4}{c}{} \\[-7pt]
\cline{2-2} \cline{3-3} \cline{4-4} \cline{5-5}
& L-CC & R-CC & L-MLO & R-MLO \\
\cline{2-2} \cline{3-3} \cline{4-4} \cline{5-5}
\end{tabular}
\caption{\label{fig:architecture}An overview of the proposed multi-view deep convolutional network. DCN refers to the convolutional network network column from Figure~\ref{fig:column}. The arrow indicates the direction of information flow.}
\end{figure}

\begin{figure*}[h!]
\centering
\begin{tabular}{c c c}
\includegraphics[trim = 0mm 0mm 0mm 0mm, scale = 0.5]{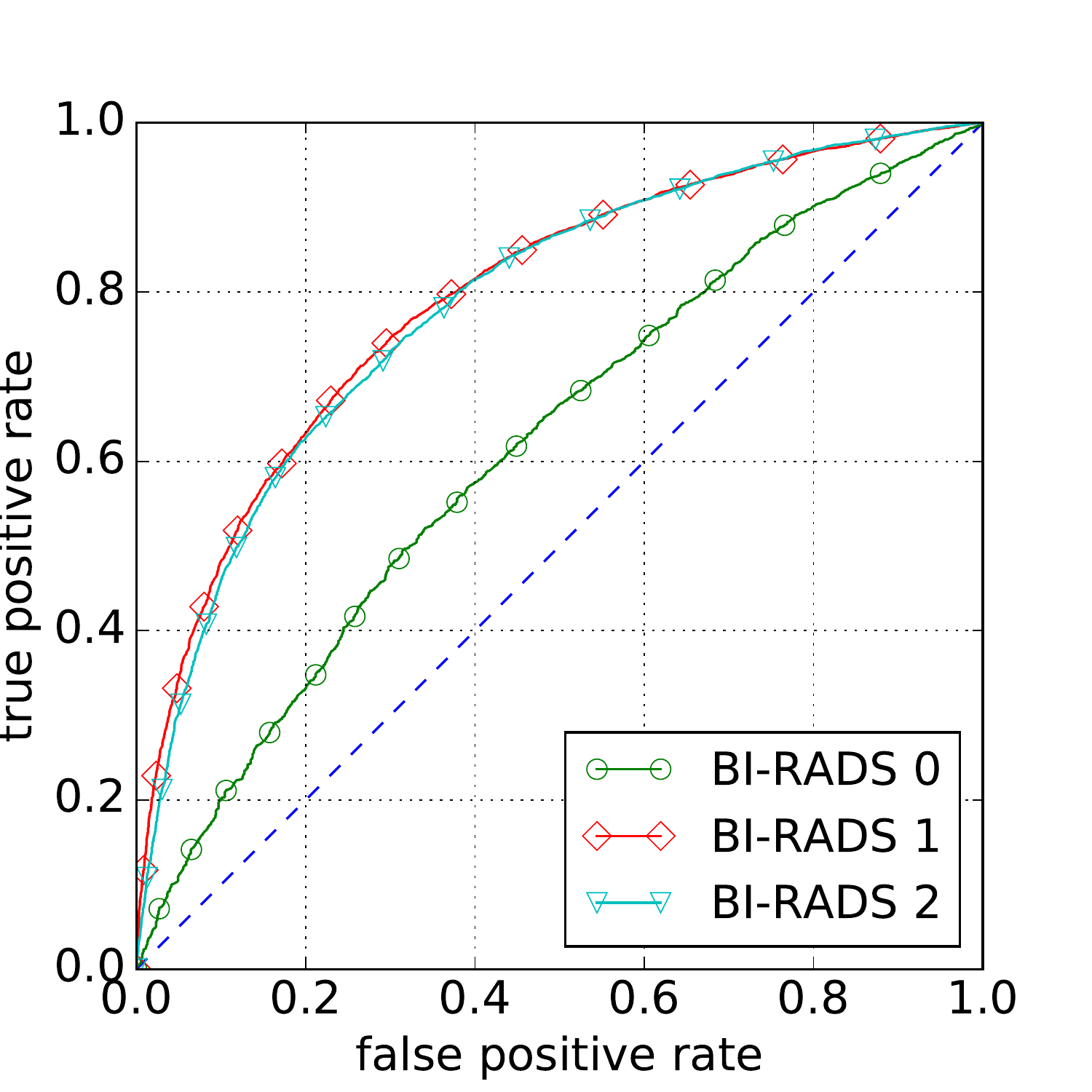} & \hspace{1cm} & \includegraphics[trim = 0mm 0mm 0mm 0mm, scale = 0.5]{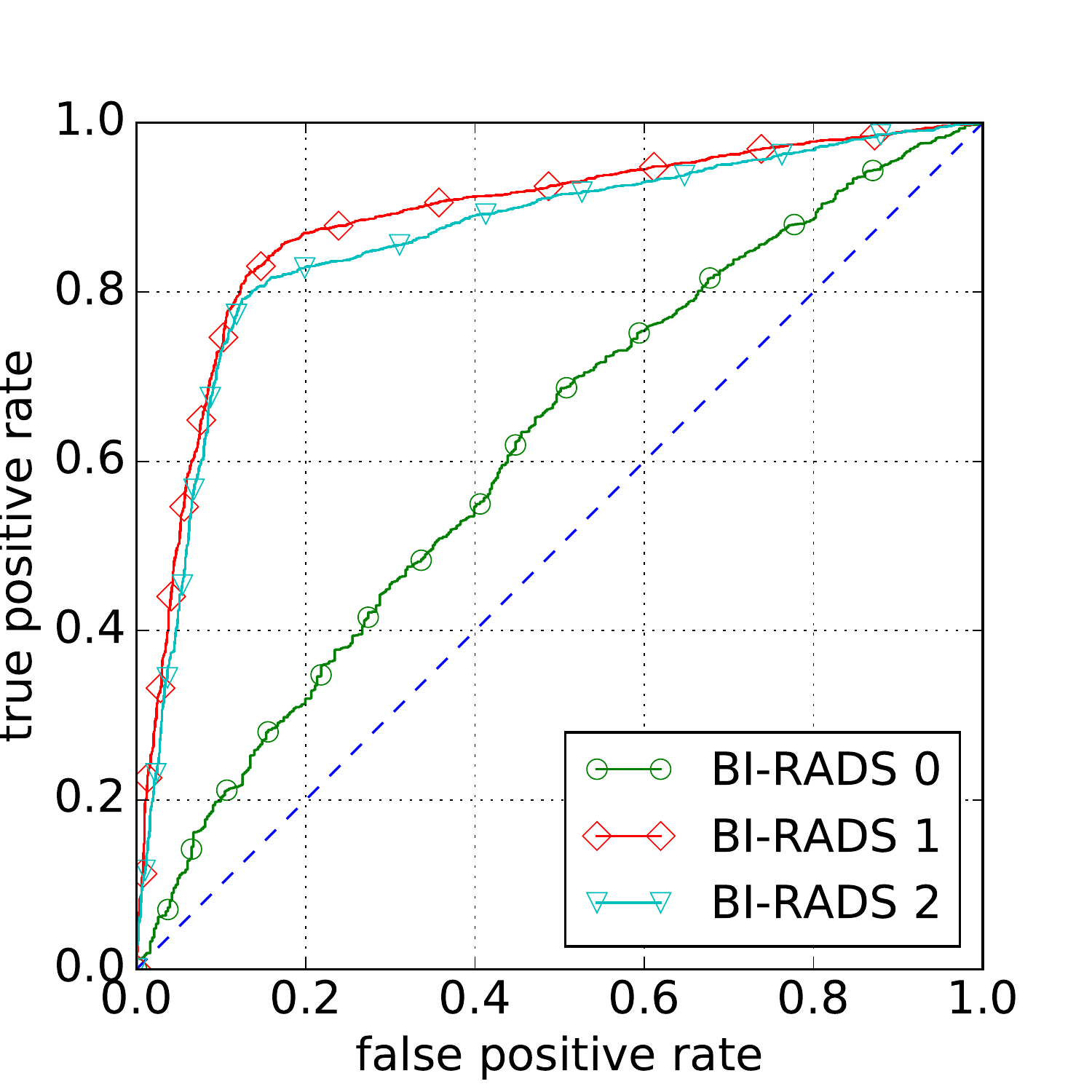}
\end{tabular}
\caption{\label{fig:ROCs}ROCs computed with all test data (left) and ROCs computed with test data which the network was confident about (right). ROCs for BI-RADS 1 and BI-RADS 2 classes improve a lot for confident examples while BI-RADS 0 remain similar.}
\end{figure*}

The overall architecture of our network is shown in Figure~\ref{fig:architecture}. Each column corresponding to a different view has an architecture described in Figure~\ref{fig:column}. We applied the rectifier function after convolutional layers. In addition to augmenting the data by cropping the images at random positions, we regularized the network in three ways. First, we tied the weights in the corresponding columns, i.e., the parameters of the columns processing L-CC and R-CC views were shared as were those of the columns processing L-MLO and R-MLO views. Second, we added Gaussian noise to the input (with the mean of zero and the standard deviation of 0.01). Third, we applied dropout (with a rate of 0.2) after the fully connected layer. We turned off the input noise and dropout during validation and testing.

The parameters of the network were initialized using the recipe of Glorot \& Bengio \cite{glorot2010understanding} and learned using the Adam algorithm \cite{kingma2015adam} with the initial learning rate of $10^{-5}$. Due to the memory limitations of our hardware, the mini-batch size was set to four. We trained the network for up to 100 epochs, which takes approximately four weeks using one NVIDIA Tesla V100 GPU. After each training epoch we computed the macAUC on the validation set. We reported the test error of the model which achieved the lowest macAUC on the validation set.

We made the code allowing to run our best network available online at \url{https://github.com/nyukat/BIRADS_classifier}.

\section{Quantitative Results Analysis}

\subsection{Effect of Scale}

First, we validated our earlier claim on the need of large-scale data for harnessing the most out of deep convolutional neural networks. We trained separate networks on the training sets of different sizes; 100\%, 50\%, 20\% and 10\%, 5\%, 2\% and 1\% of the original training set\footnote{We created the subsets of the original training set by random sampling without replacement.}. In Table~\ref{tab:size}, we observed that the classification performance improves as the number of training examples increases. This shows the importance of using a large training set. This is consistent with observations made in many other fields such as computer vision, natural language processing and speech recognition \cite{lecun2015deep}.

\begin{table}[b]
\centering
\caption{\label{tab:size} The effect of changing the fraction of the training data used. Increasing the amount of data yields better results.
}
\begin{tabular}{c | c c c c c c c}
fraction & 1\% & 2\% & 5\% & 10\% & 20\% & 50\% & 100\%\\
\hline\hline
0 vs. others & 0.541 & 0.550 & 0.559 & 0.564 & 0.570 & 0.604 & 0.618\\
1 vs. others & 0.534 & 0.631 & 0.707 & 0.738 & 0.749 & 0.774 & 0.794\\
2 vs. others & 0.537 & 0.628 & 0.715 & 0.742 & 0.752 & 0.771 & 0.787\\
\hline
macAUC & 0.537 & 0.603 & 0.660 & 0.681 & 0.690 & 0.716 & 0.733\\
HC-macAUC & 0.554 & 0.652 & 0.710 & 0.751 & 0.744 & 0.778 & 0.787\\
\end{tabular}
\end{table}

\subsection{Effect of Resolution}

We then investigated the effect of resolution of input images. Using the full training set, we trained networks with varying input resolutions; scaling both dimensions of the input by $\times$1/8, $\times$1/4 and $\times$1/2. We used bicubic interpolation to downscale the input. When the input resolution is significantly smaller than the original some convolutional layers in the later stages cannot be applied because the size of the feature maps becomes smaller than the size of a convolutional kernel. In that case, we simply skipped the remaining layers until the global average pooling. As shown in Table~\ref{tab:res}, we already saw a drop in performance when each dimension of the input was downscaled by half. Further degradation of performance was observed with more aggressive downscaling.

\begin{table}
\centering
\caption{\label{tab:res} The effect of decreasing the resolution of the image.}
\begin{tabular}{c | c c c c}
scale & $\times$1/8 & $\times$1/4 & $\times$1/2 & $\times$1\\
\hline\hline
0 vs. others & 0.587 & 0.585 & 0.611 & 0.618\\
1 vs. others & 0.718 & 0.742 & 0.779 & 0.794\\
2 vs. others & 0.729 & 0.750 & 0.777 & 0.787\\
\hline
macAUC & 0.678 & 0.692 & 0.722 & 0.733\\
HC-macAUC & 0.743 & 0.753 & 0.782 & 0.787\\
\end{tabular}
\end{table}

\subsection{Confidence}

Additionally, we checked how our model is performing for test examples depending on how confident it is about its predictions. We measure confidence of predictions in terms of the entropy of the output distribution (cf. Equation~\ref{eq:entropy}). This is how we performed the procedure allowing us to quantify this property of our model. First, we divided the exams in the validation set between the three classes. For each class separately we sorted the exams according to the entropy of predictions made by our model. Let’s define (for each class separately) $t_k$ as the threshold such that $k$ percent of the examples in the validation set have entropy (of the predictions made by our model) smaller than $t_k$. Then, for the examples from the test set (again, for each class separately) and we selected only those for which the entropy (of the prediction of our model) was lower than $t_k$. For these examples, we recomputed AUCs and macAUC. When k = 30 we call macAUC computed for this subset of data the high confidence macAUC (HC-macAUC). As shown in Table~\ref{tab:conf}, confident predictions of the proposed model are more accurate. 
This phenomenon was apparent in in all the experiments (see Table~\ref{tab:size}, Table~\ref{tab:res} and Figure~\ref{fig:ROCs}).

\begin{table}[h!]
\caption{\label{tab:conf} Average AUC (macAUC) as a function of the confidence threshold $T_{P\%}$. When $P = 30\%$, we refer to the macAUC as a high-confidence macAUC (HC-macAUC).}
\begin{center}
\begin{tabular}{c | c c c c c c}
$T_{P\%}$ & $T_{10\%}$ & $T_{20\%}$ & {\boldmath $T_{30\%}$} & $T_{50\%}$ & $T_{100\%}$\\
\hline\hline
macAUC & 0.865 & 0.827 & 0.811 & 0.781 & 0.732\\
\end{tabular}
\end{center}
\end{table}

\section{Visualization}

\begin{figure*}[h!]
\centering
\begin{minipage}{0.49\textwidth}
\centering
\includegraphics[width=\columnwidth, trim=0mm 0mm 0mm 0mm, clip]{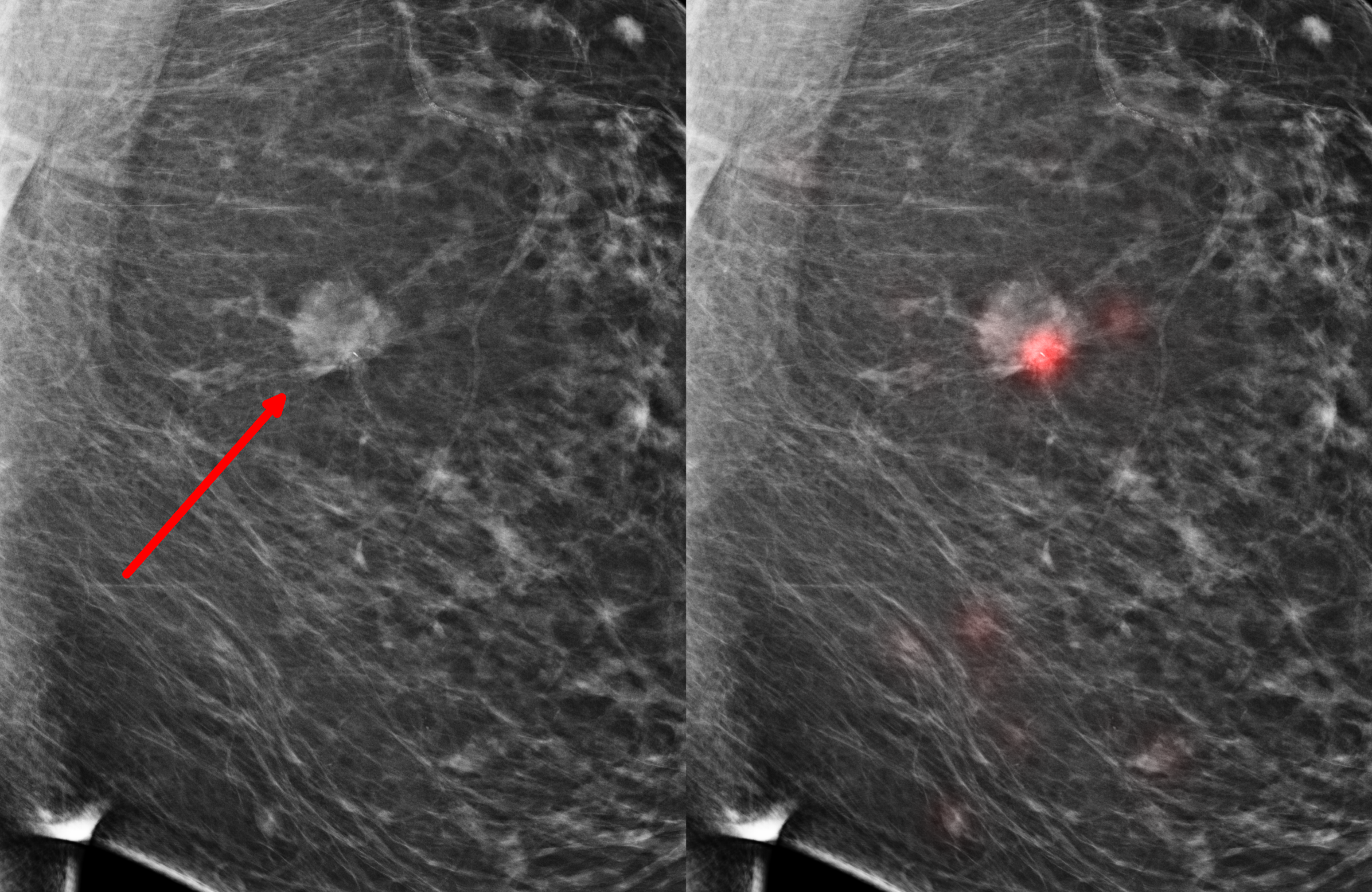}\\
\footnotesize{(a) BI-RADS 0}
\end{minipage}
\hfill
\begin{minipage}{0.49\textwidth}
\centering
\includegraphics[width=\columnwidth, trim=0mm 0mm 0mm 0mm, clip]{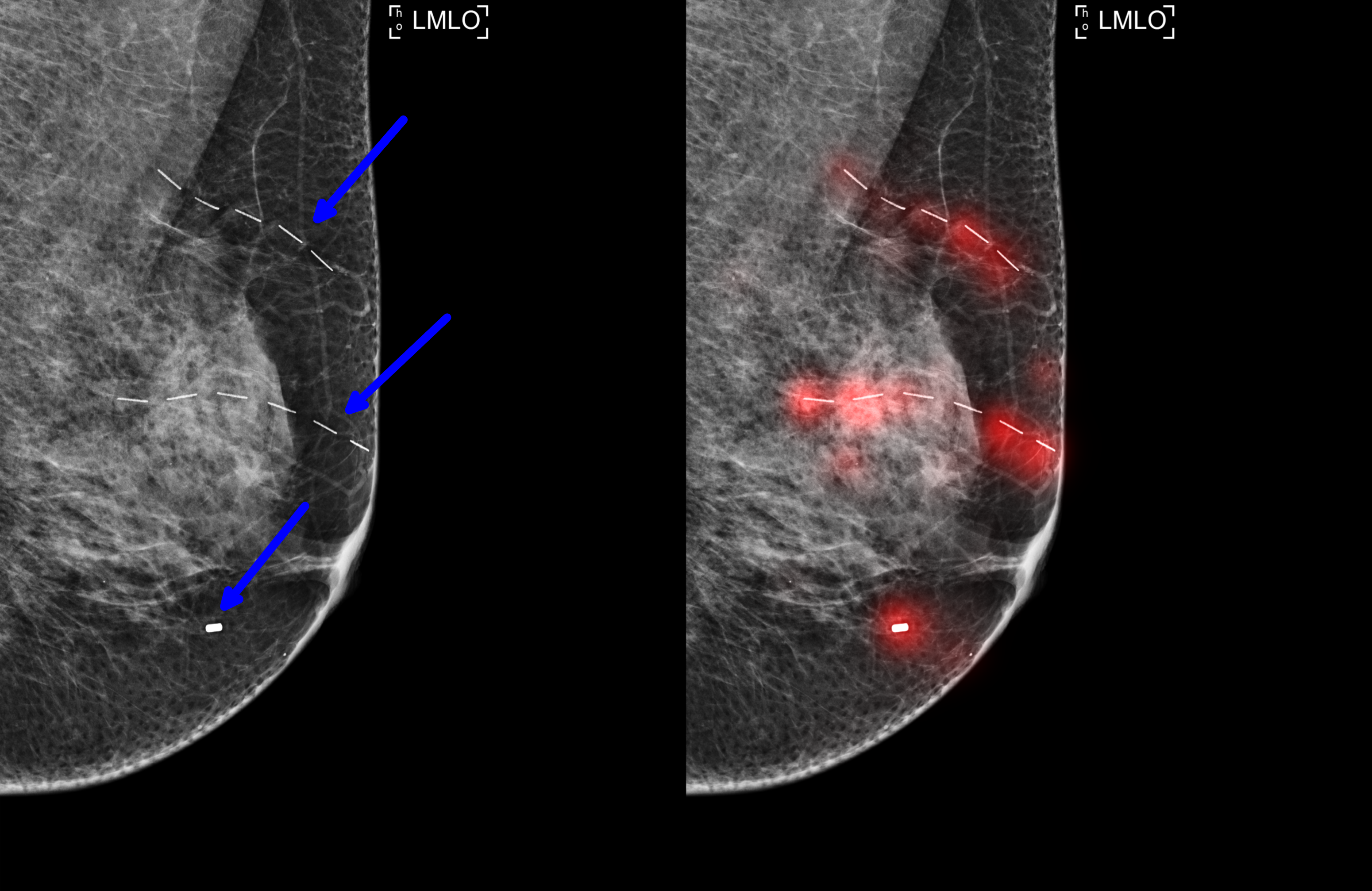}\\
\footnotesize{(b) BI-RADS 2}
\end{minipage}

\caption{Examples of visualizations of decisions made by our model. On the left in both (a) and (b) panels there are images of breast with arrows indicating possible suspicious findings. On the right in both (a) and (b) panels there are the same images as on the left in the corresponding panels with regions of the images (highlighted in red) which influence confidence of predictions of our neural network. Please note that our visualization highlights parts of the image that are relevant for all classes (BI-RADS 0, BI-RADS 1, and BI-RADS 2) and that those highlighted areas include locations indicated in the images on the left. Panel (a) shows the right breast of a 61 years old patient who was assigned BI-RADS 0 in her screening mammography. Biopsy confirmed that the finding indicated in the image by the right arrow was invasive ductal carcinoma. Panel (b) shows the left breast of a 62 years old patient. She had a prior breast surgery and a biopsy marker in one of her breasts as indicated by the blue arrows. Our neural network correctly and confidently predicted that the artifacts were benign and indicated scar markers and a biopsy clip.}
\label{fig:per-case}
\end{figure*}

A flip side of high effectiveness of a deep convolutional neural network is the difficulty in interpreting its internal processing.
Only recently there have been some efforts on visualizing deep convolutional neural networks for computer vision \cite{zeiler2014visualizing,yosinski2015understanding}. These recent approaches, however, are not computationally efficient and are not easy to apply to medical images for a number of reasons, including the need for training with a large data set \cite{zeiler2014visualizing} and the availability of good image statistics \cite{yosinski2015understanding}. Instead, we propose a simpler visualization technique in this paper that does not require any further training.

We look at the sensitivity of the network's output to the perturbation of each input pixel. The network outputs the conditional distribution over all the categories, and we can measure the entropy (or confidence) of the predictive distribution $\mathcal{H}(y|\mathbf{x})$. We can use standard backpropagation to compute $\left| \frac{\partial \mathcal{H}}{\partial \mathbf{x}_{ij}^v}\right|$ for the pixel $(i,j)$ of the $v$-th view. Those input pixels that influence the confidence of the network will have high values, and those that do not contribute much will have low values ($\approx 0$). We show two examples of such visualization for patients which were confirmed by a follow-up examination to have breast cancer in Figure~\ref{fig:per-case}.

\section{Reader study}

To understand the limit of performance possible to achieve on this task, we conducted a reader study with four human experts, who all were doctors experienced in reading breast cancer screening exams. The experts were all shown the same 500 exams randomly drawn from the test set, each with at least four images corresponding to the standard views used in screening mammography. For each exam, they were asked to indicate the most likely BI-RADS label according to their judgement. We first measured agreement between the radiologists themselves and between the radiologists and the labels in the data. The results are shown in Table~\ref{tab:reader_study}. We can clearly observe that the agreement between different radiologist as well as between radiologists and the labels in the data is low. To obtain probabilistic predictions from this group of experts, we represented their classifications as one-hot vectors and averaged them for each exam. On the random subset of data used in our reader study (which turned out to be a difficult subset, cf. Table~\ref{tab:size}) such a committee of radiologists achieved the macUAC of 0.704, while our model achieved the macUAC of 0.688. We conclude from these results that predicting BI-RADS without prior exams and information about the patient is very difficult even for well-trained human experts. Our neural network is already performing well in comparison. It is interesting to note that our model is clearly worse than the committee of the radiologists in recognizing BI-RADS 0 and clearly better in recognizing BI-RADS 2. We also evaluated an ensemble, created by equally weighting predictions of the committee of radiologists and our neural network (Table~\ref{tab:reader_study}). For each of the BI-RADS categories, the ensemble was at least as good as any of its two base elements.

\begin{table}[htb!]
\centering
\caption{\label{tab:reader_study} Results of our reader study comparing accuracies obtained by the committee of radiologists, our neural network (MV-DCN) and an ensemble of the two.}
\begin{tabular}{c | K{1.8cm} | K{1.8cm} | K{1.8cm}}
& radiologists & MV-DCN & radiologists + MV-DCN\\
\hline\hline
0 vs. others & 0.650 & 0.547 & 0.653\\
1 vs. others & 0.765 & 0.757 & 0.792\\
2 vs. others & 0.699 & 0.759 & 0.759\\
\hline
macAUC & 0.704 & 0.688 & 0.735\\
\end{tabular}
\end{table}

\begin{table}[htb!]
\caption{Agreement (Cohen's kappa) in BI-RADS categorization between different radiologists (R1, R2, R3, R4) and labels in the data set (L).}
\label{tab:consistency_2}
\vspace{-3mm}
\begin{center}
\begin{tabu}{| c | [2pt]c | c | c | c | c |}
\cline{2-6}
\multicolumn{1}{c|}{} & L & R1 & R2 & R3 & R4\\ \tabucline[2pt]{2-6} \hline
L & \cellcolor{gray!25} & 0.29 & 0.24 & 0.24 & 0.26\\ \hline
R1 & \cellcolor{gray!25} & \cellcolor{gray!25} & 0.29 & 0.34 & 0.35\\ \hline
R2 & \cellcolor{gray!25} & \cellcolor{gray!25} & \cellcolor{gray!25} & 0.48 & 0.45 \\ \hline
R3 & \cellcolor{gray!25} & \cellcolor{gray!25} & \cellcolor{gray!25} & \cellcolor{gray!25} & 0.50\\ \hline
R4 & \cellcolor{gray!25} & \cellcolor{gray!25} & \cellcolor{gray!25} & \cellcolor{gray!25} & \cellcolor{gray!25}\\ \hline
\end{tabu}
\end{center}
\vspace{-7mm}
\end{table}

\section{Conclusions}

In this paper we have made a first step towards end-to-end large scale training of multi-view deep convolutional networks for breast cancer screening. We have shown experimentally that it is essential to keep the images at high-resolution. We expect this to hold for other learning tasks with medical images where fine details determine the outcome. We also demonstrated it is necessary to use a large number of exams. Although we used the largest breast cancer screening data set ever reported in literature, the performance of our model has not saturated and is expected to improve with more data.

Our network's performance, just like performance of the doctors participating in our reader study, was lowest on differentiating BI-RADS 0 from the other classes. Doctors often disagree on how a particular exam should be classified \cite{RN40} and in fact, less than 1\% of the screening population has cancer \cite{RN38,RN37}. We expect that this problem can be alleviated by using instead the information on whether a person actually went on to develop breast cancer in the future as a label.

It is also worth noting that, because of limited computational resources, we had to heavily rely on our experience in the choice of learning hyperparameters. We did not perform a systematic search for optimal hyperparameters, which often has a great impact on the performance of a neural network in limited data scenarios \cite{algorithms_for_hyper,snoek2015scalable}. The methods we used in this work are powerful and our results can be improved simply by the means of applying more computational resources without significantly changing the methodology.

\section*{Acknowledgments} We would like to thank Jure \v{Z}bontar, Yann LeCun, Pablo Sprechmann, Cem Deniz, Jingyi Su and Masha Zorin for insightful comments on this work, as well as Jason Phang and Jungkyu Park for creating a PyTorch version of the code. We would also like to thank Joe Katsnelson and Mario Videna for their efforts in supporting our computing environment.

\bibliography{bibtex/bib/IEEEexample.bib}{}
\bibliographystyle{IEEEtran}

\end{document}